\documentclass{article}
\usepackage[nonatbib,final]{neurips_2024}

\usepackage[utf8]{inputenc} 
\usepackage[T1]{fontenc}    
\usepackage[pagebackref=true,breaklinks=true,colorlinks,bookmarks=false]{hyperref}
\usepackage{url}            
\usepackage{booktabs}       
\usepackage{amsfonts}       
\usepackage{nicefrac}       
\usepackage{microtype}      
\usepackage{xcolor}         
\usepackage{graphicx}
\usepackage{kotex}
\usepackage{multirow}
\usepackage{xspace}
\usepackage{amsthm}
\usepackage{algorithm}
\usepackage{algorithmic}
\usepackage{enumitem}
\usepackage{wrapfig}
\usepackage{subcaption}
\setlist[itemize]{leftmargin=20pt, rightmargin=0pt}

\newcommand{\rb}{\rotatebox{90}}

\makeatletter
\DeclareRobustCommand\onedot{\futurelet\@let@token\@onedot}
\def\@onedot{\ifx\@let@token.\else.\null\fi\xspace}

\def\eg{\emph{e.g}\onedot} 
\def\ie{\emph{i.e}\onedot}

\makeatother

\newcommand\framework{{MapUnveiler}}

\title{Unveiling the Hidden: \\Online Vectorized HD Map Construction with \\Clip-Level Token Interaction and Propagation}

\author{
  Nayeon Kim\thanks{Equal contribution.} \quad\quad Hongje Seong$^*$ \quad\quad Daehyun Ji \quad\quad Sujin Jang\thanks{Corresponding author.} \\
  Samsung Advanced Institute of Technology (SAIT) \\
  \texttt{\{nayeon.kim,~hongje.seong,~derek.ji,~s.steve.jang\}@samsung.com} \\
}

\begin{document}

\maketitle

\begin{abstract}
Predicting and constructing road geometric information (\eg, lane lines, road markers) is a crucial task for safe autonomous driving, while such static map elements can be repeatedly occluded by various dynamic objects on the road.
Recent studies have shown significantly improved vectorized high-definition (HD) map construction performance, but there has been insufficient investigation of temporal information across adjacent input frames (\ie, clips), which may lead to inconsistent and suboptimal prediction results.
To tackle this, we introduce a novel paradigm of clip-level vectorized HD map construction, \textbf{\textit{\framework}}, which explicitly unveils the occluded map elements within a clip input by relating dense image representations with efficient clip tokens.
Additionally, \framework{} associates inter-clip information through clip token propagation, effectively utilizing long-term temporal map information.
\framework{} runs efficiently with the proposed clip-level pipeline by avoiding redundant computation with temporal stride while building a global map relationship.
Our extensive experiments demonstrate that \framework{} achieves state-of-the-art performance on both the nuScenes and Argoverse2 benchmark datasets.
We also showcase that \framework{} significantly
outperforms state-of-the-art approaches in a challenging setting, achieving 
+10.7\% mAP improvement in heavily occluded driving road scenes.
The project page can be found at \url{https://mapunveiler.github.io}.
 
\end{abstract}
\section{Introduction}
\label{section1}
Vectorized HD map construction (\textit{\textbf{VHC}}) is a task of predicting instance-wise vectorized representations of map elements (\eg, pedestrian crossings, lane dividers, road boundaries).
Such static map elements are crucial information for self-driving vehicles, including applications like lane keeping~\cite{amditis2010situation,chen2017end}, path planning~\cite{meyer2003map,liu2017path,hu2023planning}, and trajectory prediction~\cite{narayanan2021divide,wang2022ltp,gu2024producing}.
Prior approaches to constructing dense and high-quality HD maps typically rely on SLAM-based offline methods~(\eg,~\cite{zhang2014loam,shan2018lego,shan2020lio}).
Such an offline method generally involves a series of steps including feature extraction and selection (\eg, edge, plane), odometry estimation via feature matching, and mapping.
However, these processes involve complicated and computationally burdensome tasks, limiting their use to offline applications.

\begin{figure}
    \centering
    \includegraphics[width=.9\linewidth]{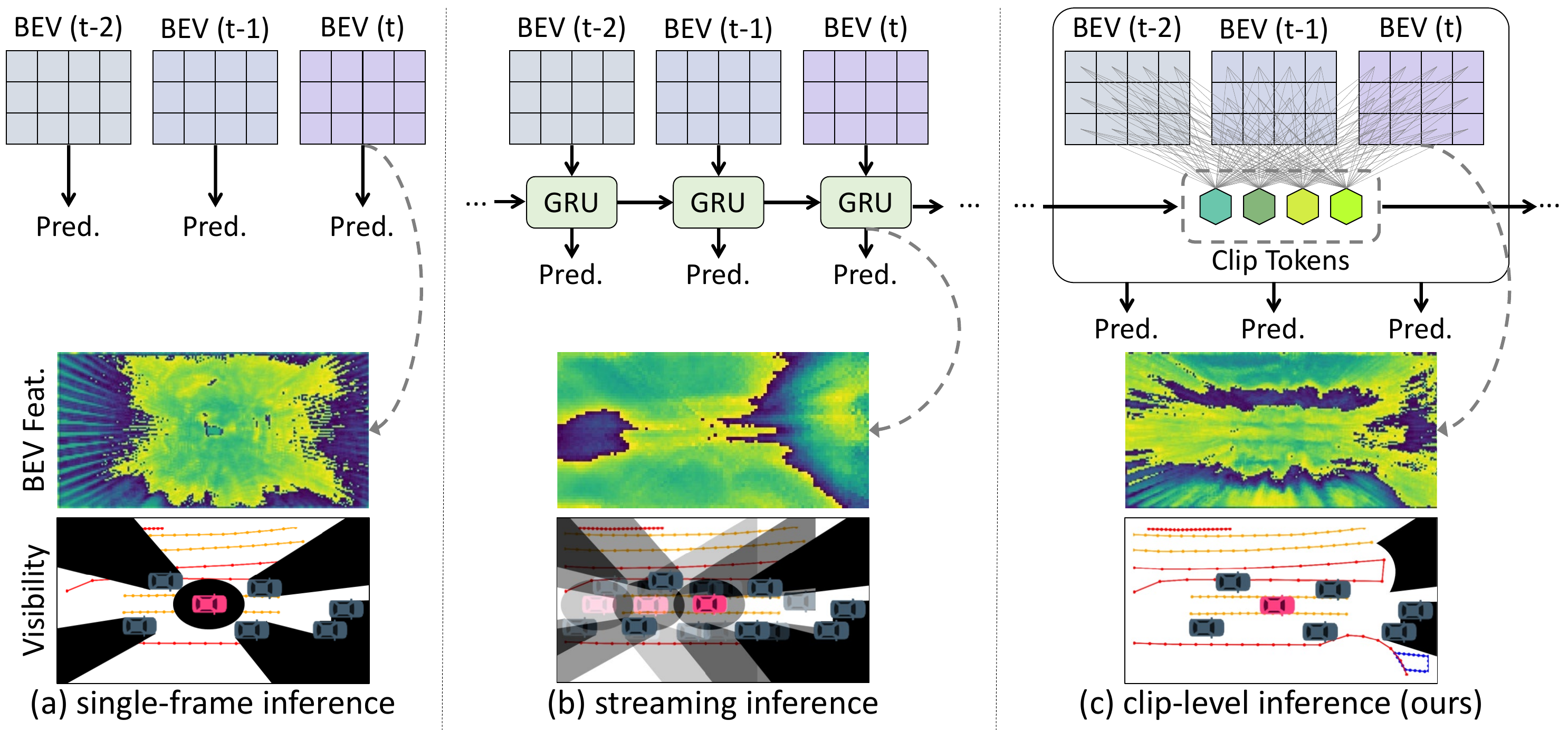}
    \caption{
    (a) Existing approaches relying on single-frame inference cannot capture the entire map information in the BEV features~\cite{liu2023vectormapnet,liao2023maptr,liao2023maptrv2}.
    (b) Recent alternatives explore temporal information via streaming, but they cannot address the inherent nature of maps and propagate noise from previous timestamps.
    (c) We directly unveil hidden maps in BEV features by interacting with clip tokens that contain high-level map information.
    We visualize BEV features by 1D PCA projection. The BEV features are extracted from (a)~MapTRv2~\cite{liao2023maptrv2}, (b)~StreamMapNet~\cite{yuan2024streammapnet}, and (c)~our \framework{}.
    }
    \label{fig:fig1}
    \vspace{-4mm}
\end{figure}

More recently, camera-based multi-view VHC has been actively investigated as a cost-efficient and real-time alternative to existing expensive offline approaches.
Current works on camera-based VHC typically aim to extract unified 3D Bird's Eye View (\textit{\textbf{BEV}}) features that cover the surrounding environment of the ego-vehicle~\cite{li2022hdmapnet,liao2023maptr,liao2023maptrv2}, relying on various Perspective View (PV) to BEV transformation methods~\cite{zhou2022cross,philion2020lift,li2022bevformer,chen2022efficient}.
Subsequently, a task-specific head follows to decode and predict map elements from the extracted BEV features.
Despite significant progress, prior works still suffer from frequently occluded map elements caused by dynamic foreground objects such as vehicles and pedestrians, as described in Fig.~\ref{fig:fig1}-(a).
Moreover, the prediction performance degrades when applied to a larger perception range.
To address such issues, prior works try to leverage temporal information extracted from a stream of preceding feature frames~\cite{yuan2024streammapnet,wang2024stream}.
While such approaches have achieved improved \textit{online} prediction performance, they still do not fully leverage the potential of temporal information across a longer range of frames.
In particular, prior works do not consider the cumulative impacts of occluded map elements, leading to noisy BEV feature generation, as shown in Fig.~\ref{fig:fig1}-(b) and Tab.~\ref{tab:occlusion}.
Moreover, such cumulative flaws can ultimately lead to degraded performance in longer perception ranges (see Tab.~\ref{tab:comparison_nuscenes}), which is crucial for safety-critical autonomous driving.
On the other hand, the key idea of conventional SLAM-based offline methods is to stack and associate static features collected over a longer window of frames (\ie, mapping features).
This \textit{offline} mapping strategy effectively addresses occlusion issues by leveraging diverse views of the map elements across multiple frames, while it requires manual human annotation and complex pipelines.

Based on these insights, we introduce a novel clip-level construction framework, \textit{\textbf{\framework{}}}, which incorporates the effective offline mapping strategy into state-of-the-art online VHC approaches.
In contrast to the direct \textit{global feature mapping} used in offline methods, our \framework{} collects \textit{clip-level} temporal map information and learns differentiable associations for efficient online inference.
We generate compact clip tokens consisting of temporal map information within a clip input and update BEV features with these tokens to unveil hidden map elements that are visible in certain frames, as shown in Fig.~\ref{fig:fig1}-(c).
Subsequently, we transfer the inter-clip tokens to the next clip's BEV features to facilitate establishment the long-term intra-clip associations among map elements.

As a result, \framework{} outperforms state-of-the-art methods on two widely recognized benchmarks: +1.3\% mAP on nuScenes~\cite{caesar2020nuscenes} and +0.9\% mAP on Argoverse2~\cite{Argoverse2}.
We also observe a marginal increase in computational burden, as we exploit the benefits of the clip-level inference strategy, which allows an efficient inference of multi-frame inputs.
In summary, our contributions include:
\vspace{-\topsep}
\vspace{-1mm}
\begin{itemize}
\item We propose \framework{}, an online VHC model that incorporates the offline mapping strategy and unveils the hidden maps by interacting multiple dense BEV features with compact tokens.
\item We introduce the \textit{clip-level} pipeline to infer \framework{} online by mapping within a clip set of BEV features and propagating the map information to subsequent timestamps, thereby building a global map efficiently.
\item Our method significantly improves a frame-level model on longer perception range settings (58.6\%~$\rightarrow$~68.7\%) and heavy occlusion splits (53.1\%~$\rightarrow$~63.8\%), and achieves state-of-the-art performance on two standard VHC benchmarks with marginal extra computations (15.6~FPS~$\rightarrow$~12.7~FPS).
\end{itemize}

\begin{figure}
    \centering
    \includegraphics[width=1\linewidth]{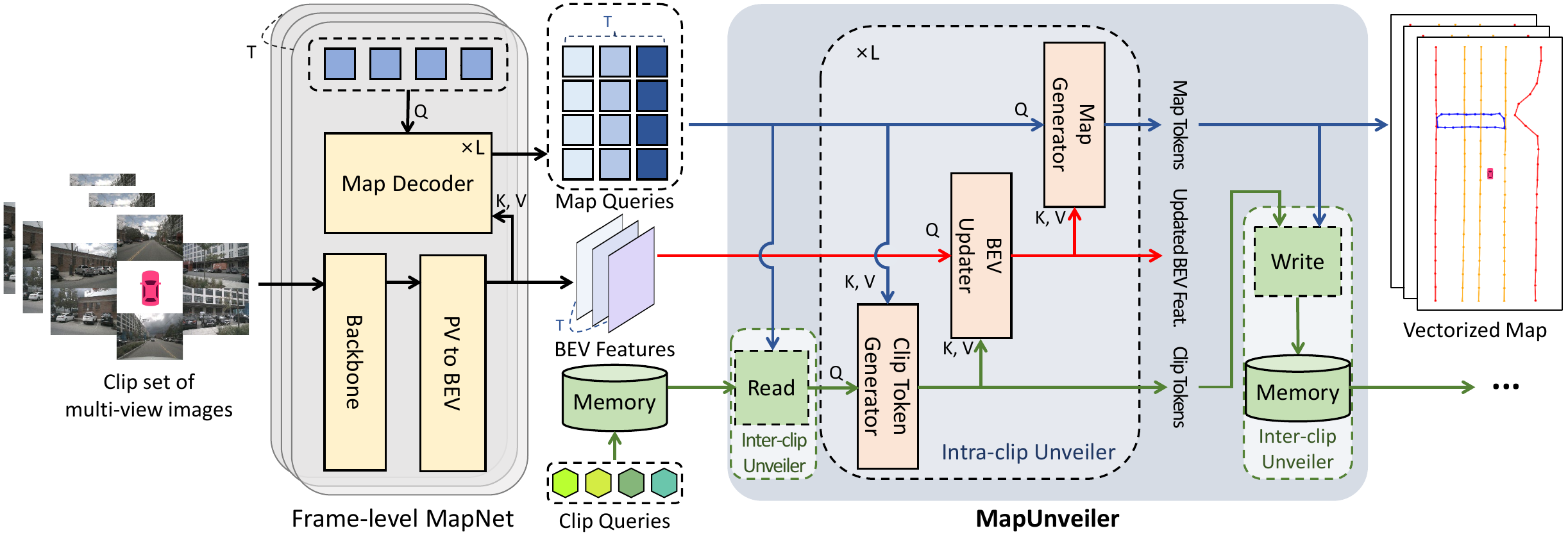}
    \caption{
    Our framework takes clip-level multi-view images and outputs clip-level vectorized HD maps.
    All components in the frame-level MapNet (\ie, Backbone, PV to BEV, Map Decoder) are adopted from MapTRv2~\cite{liao2023maptrv2}.
    The frame-level MapNet extracts map queries and BEV features independently at each frame.
    \framework{} generates compact clip tokens that contain clip-level temporal map information and directly interact with dense BEV features.
    With the updated BEV features, we construct high-quality clip-level vectorized maps.
    The generated map tokens and clip tokens are then written to memory.
    }
    \label{fig:overview}
    \vspace{-2mm}
\end{figure}

\section{Related Work}
\label{section2}
\paragraph{Multi-View HD Map Construction.}
SLAM (Simultaneous Localization and Mapping)~\cite{durrant2006simultaneous} has been a central technique for constructing accurate HD maps~\cite{zhang2014loam,shan2018lego,shan2020lio}.
However, these methods require memory-intensive, complex pipelines for global mapping of geometric features, and are therefore typically executed offline.
Recently, deep learning approaches have emerged as an appealing alternative to those expensive offline methods, enabling online HD map construction using cost-efficient multi-view camera sensors.
The perspective-view (PV) to bird's-eye-view (BEV) transformation methods~\cite{zhou2022cross,philion2020lift,li2022bevformer,chen2022efficient} enable the generation of 3D features from the surrounding environment of the ego-vehicles using camera sensors, even in the absence of precise spatial cues.
BEVFormer~\cite{li2022bevformer} utilizes the deformable attention mechanism~\cite{zhu2021deformable} to extract BEV features and predict rasterized semantic maps.
However, it cannot generate instance-wise representation of map elements.
To address this, HDMapNet~\cite{li2022hdmapnet} introduces a heuristic method to group pixel-level semantic maps into a vectorized representation.
Similarly, VectorMapNet~\cite{liu2023vectormapnet} proposes an end-to-end learning approach to predicting vectorized map representations.
Although such methods have demonstrated notable prediction performance in single-frame inference, they do not consider the temporal information from multi-frame inputs.
More recently, StreamMapNet~\cite{yuan2024streammapnet} and SQD-MapNet~\cite{wang2024stream} have proposed a streaming feature paradigm~\cite{wang2023exploring}, which aims to leverage temporal information for improved temporal consistency across predictions.
However, these methods propagate dense BEV features directly, incorporating map information from previous frames that may have been occluded and undetected, resulting in the accumulation of noise, as shown in Fig.~\ref{fig:fig1}-(b).
To address this issue, we propose an end-to-end clip token learning approach that combines offline mapping techniques with online strategies, aiming for high performance and computational efficiency.

\paragraph{Temporal Token Learning.}
With the rapid development of transformers~\cite{vaswani2017attention}, there has been significant interest in efficient token learning alongside dense features, \eg, CNN representations.
In particular, temporal token learning has emerged as an attractive alternative to memory-intensive spatio-temporal dense CNN representations~\cite{yang2019video,oh2019video}.
VisTR~\cite{wang2021end} extends DETR~\cite{carion2020end} into the 3D domain to extract spatio-temporal instance tokens that can be directly used for instance segmentation.
IFC~\cite{hwang2021video} proposes an efficient spatio-temporal token communication method, which replaces the heavy interactions within dense CNN features.
VITA~\cite{heo2022vita} learns efficient video tokens from frame-level instance tokens without dense CNN features.
Cutie~\cite{cheng2023putting} updates CNN representations with tokens to avoid spatio-temporal dense matching.
TTM~\cite{ryoo2023token} introduces an efficient long-term memory mechanism by summarizing tokens into memory rather than stacking~\cite{burtsev2020memory,le2019learning,rae2020compressive} or recurrence~\cite{hochreiter1997long,chung2014empirical}.
While all the aforementioned approaches were designed to handle foreground instances, we discover the potential of token learning to construct background maps.
By learning compact tokens and interacting with dense BEV features, we impose traditional mapping into online VHC model and enable online running.

\section{Method}
\label{section3}
\subsection{Overview}
\label{subsection3:Overview}
We present the overall architecture of \framework{} in Fig.~\ref{fig:overview}.
Given a set of synchronized multi-view images (\ie, clip inputs), our model sequentially construct clip-level vectorized HD maps.
We first extract frame-level BEV features and map queries, which are then used as inputs to \framework{} module.
We employ memory tokens, which are written from the previous clip and facilitate the establishment of long-term temporal relationships.
From the memory tokens and map queries, we generate clip tokens that embed temporal map element cues in a compact feature space.
This is the first step to understand clip-level map information.
We then update BEV features with clip tokens, which is the core step of unveiling hidden maps.
Using the updated (unveiled) BEV features, we extract map tokens and construct clip-level vectorized HD maps.
After a clip-level inference, we generate new memory tokens using clip tokens, map tokens, and the current memory tokens.
The new memory tokens are used for providing temporal cues for the subsequent clip-level inference.
Since we opt for a clip-level pipeline, \framework{} efficiently infers with a temporal stride $S$, performing clip-level inference only $N_{T}/S$ times for a sequence of $N_{T}$ frames.
In the following subsections, we detail each module in the proposed framework.

\subsection{Frame-level MapNet}
\label{subsection3:Frame-level}
We adopt MapTRv2~\cite{liao2023maptrv2} as our frame-level MapNet architecture to extract a clip set of BEV features and map queries from synchronized multi-view images.
We extract perspective view (PV) image features using a backbone network, then these PV image features are transformed into BEV features through a PV-to-BEV module.
Following the setup of MapTRv2, we adopt ResNet50~\cite{he2016deep} and Lift, splat, shoot (LSS)~\cite{philion2020lift}-based BEV feature pooling~\cite{huang2022bevpoolv2} for our backbone and PV-to-BEV module, respectively.
These BEV features are utilized for querying maps in the map decoder.
With their BEV features, the map decoder outputs frame-level map queries which can be directly used for constructing vectorized HD maps.
Finally, the frame-level MapNet outputs BEV features and map queries, which are the results from the PV-to-BEV module and the map decoder, respectively.
BEV features represent rasterized map features, whereas map queries embed vectorized map information; thus, we can directly construct vectorized HD maps using the map queries.

\begin{figure}
    \centering
    \includegraphics[width=1\linewidth]{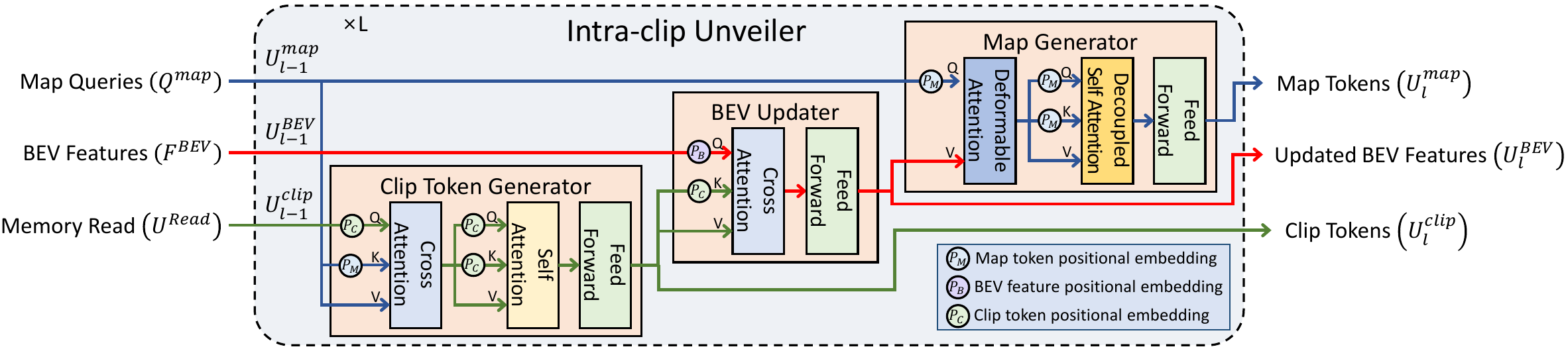}
    \caption{
    A detailed implementation of Intra-clip Unveiler.
    We use \textcolor{blue}{blue}, \textcolor{red}{red}, and \textcolor{green}{green} arrows to indicate the flows of map tokens, BEV features, and clip tokens, respectively.
    In each attention and feed forward layer, standard layer normalization, dropout, and residual connections are followed.
    }
    \label{fig:intra_clip_unveiler}
    \vspace{-2mm}
\end{figure}

\subsection{\framework{}}
\label{subsection3:Map}
\framework{} is a novel framework designed to unveil invisible map information that cannot be captured by frame-level BEV features alone.
To avoid heavy computations, we adopted a clip-level inference scheme with temporal window (clip length) $T$ and stride $S$.
A detailed explanation of the inference scheme with the temporal window $T$ and stride $S$ is provided in Appendix (see Sec.~\ref{subsubsectionA:definition_inference_scheme}).
Our \framework{} consists of two main components: (1) Intra-clip Unveiler and (2) Inter-clip Unveiler.
For each clip-level pipeline, our Intra-clip Unveiler generates vectorized maps for $T$ frames.
The Inter-clip Unveiler then writes memory tokens with the tokens generated in Intra-clip Unveiler to build global relationships.

\subsubsection{Intra-clip Unveiler}
\label{subsubsection3:Intra-clip}
Intra-clip Unveiler is composed of a sequence of $L$ layers.
It initially takes a clip set of frame-level map queries \(Q^{map}\), BEV features \(F^{BEV}\), and memory read \(U^{Read}\) (read at Inter-clip Unveiler, detailed in Sec.~\ref{subsubsection3:Inter-clip}).
In the first step, compact clip tokens are created by the clip token generator.
The BEV updater then unveils hidden maps in the BEV features with the clip tokens.
Finally, map generator outputs clip-level map tokens from the updated BEV features.
The map tokens are directly used for constructing vectorized HD maps with perception heads.
We illustrate the Intra-clip Unveiler in Fig.~\ref{fig:intra_clip_unveiler}.
In the followings, we describe the detailed implementation of each module.

\paragraph{Clip Token Generator.}
Clip token generator yields clip tokens \(U_{l}^{clip} \in \ \mathbb{R} ^{N_{c} \times C}\) from frame-level map queries \(Q^{map} \ \in \ \mathbb{R}^{T\times N_{i} \times N_{p} \times C}\), where $N_{c}$, $N_{i}$, and $N_{p}$ denote the clip token size, number of predicted map element, and number of points per map element, respectively. To globally gather intra-clip map features, we opt for a naive cross-attention~\cite{vaswani2017attention}. Through this step, we obtain compact clip-level map representations, enabling efficient intra-clip communication with small-scale features.

\paragraph{BEV Updater.}
The second step is the BEV Updater, which updates bev features \(F^{BEV}\ \in \ \mathbb{R}^{T\times H\times W\times C}\) with the clip tokens \(U_{l}^{clip}\) to unveil the hidden map element information.
In cross-attention, query is derived from the bev features  \(F^{BEV}\), and the key and value are derived from the clip token \(U_{l}^{clip}\). 
The output of this step is robustly updated bev features \(U_{l}^{BEV}\in \ \mathbb{R}^{T\times H\times W\times C}\) enhanced via clip tokens for hidden areas relative to the original bev features. 
The main idea of BEV Updater is to avoid heavy computation in spatio-temporal cross attention. To achieve this, we do not directly communicate intra-clip BEV features, but instead decouple the spatial BEV features and the temporal clip tokens. We then update the spatial BEV features with compact temporal clip tokens, effectively communicating spatio-temporal information with reasonable computational costs.
The updated bev features \(U_{l}^{BEV}\) are used as value features in the next step.

\paragraph{Map Generator.}
The last step is the Map Generator, which generates map tokens \(U_{l}^{map}\in \ \mathbb{R}^{T\times N_{i} \times N_{p} \times C}\) using the updated BEV features \(U_{l}^{BEV}\) created in the previous step. The objective of this step is to generate a refined version of frame-level map queries.
As illustrated in Fig.~\ref{fig:intra_clip_unveiler}, the map generator uses deformable attention~\cite{zhu2021deformable} and decoupled self-attention~\cite{liao2023maptr} mechanisms, following~\cite{liao2023maptr}.
In deformable attention, query is derived from the map queries \(Q^{map}\), and the value is derived from the updated BEV features \(U_{l}^{BEV}\).
Since the updated BEV features are spatio-temporally communicated, we directly extract map tokens. Each map token represents a vectorized map element through a 2-layer Multi-Layer Perceptron (MLP).
The map tokens \(U_{l}^{map}\) are written to the memory of the Inter-clip Unveiler, and when the map tokens \(U_{l}^{map}\) of the $L$-th layer pass through the prediction head, vectorized maps are generated.

\subsubsection{Inter-clip Unveiler}
\label{subsubsection3:Inter-clip}
Inter-clip Unveiler propagates the tokens from previous clip input to the next one, thereby preserving the dense temporal information from the prior frames.
As shown in Fig.~\ref{fig:overview}, Inter-clip Unveiler writes map tokens \(U_{l}^{map}\) and clip tokens \(U_{l}^{clip}\) from the Intra-clip Unveiler to the memory.
Here, we adopt token turning machine (TTM)~\cite{ryoo2023token} to efficiently manage the long-term map information.
In the followings, we describe the detailed implementation of read and write.

\paragraph{Read.}
We generate compact tokens that contain a global map information by reading from memory tokens and map queries.
Following TTM~\cite{ryoo2023token}, we read with the token summarizer~\cite{ryoo2021tokenlearner} which efficiently selects informative tokens from inputs as follows:
\begin{equation}
U^{read}=Read(U_{t-2S:t-S}^{memory}, Q^{map})=S_{N_c}([U_{t-2S:t-S}^{memory}|| Q^{map}]),
\end{equation}
where $[U_{t-2S:t-S}^{memory}|| Q^{map}]$ denotes the concatenation of two elements, and $U_{t-2S:t-S}^{memory}$ denotes memory tokens for a clip.
We employ the location-based memory addressing used in~\cite{ryoo2023token} utilizing the positional embedding (detailed in Section~\ref{subsubsection4:positional embedding}).
Note that the memory is not available in the first clip-level pipeline.
Therefore, we initially write the memory token from learnable clip embeddings.

\paragraph{Write.}
We employ the write operation with the same token summarizer~\cite{ryoo2021tokenlearner} that is used in~\cite{ryoo2023token}.
The new memory \(U_{t-S:t}^{memory}\in \ \mathbb{R}^{M\times C}\) is generated by summarizing the clip tokens \(U_{L}^{clip}\), map tokens \(U_{L}^{map}\), and old memory \(U_{t-2S:t-S}^{memory}\) as follows:
\begin{equation}
U_{t-S:t}^{memory}=Write(U_{L}^{clip}, U_{L}^{map}, U_{t-2S:t-S}^{memory})=S_M([U_{L}^{clip}|| U_{L}^{map} ||U_{t-2S:t-S}^{memory}]),
\end{equation}
where $M$ denotes the memory token size.
The newly generated memory through the write operation is used in the read operation of the first layer of the Intra-clip Unveiler in the subsequential per-clip process.
If the tokens within the memory are not re-selected in the subsequent steps, it will be removed from the memory, and the selection mechanism will be determined through the learning.
We employ the same memory addressing method used in the read operation.
The write operation is applied in the last layer of the Intra-clip Unveiler, generating new memory that preserves the information of the clip tokens and map tokens.

\subsubsection{Positional Embedding}
\label{subsubsection4:positional embedding}
While the standard transformer structure is permutation-invariant, we require position information added with temporal information to predict map elements at the clip-level.
For the BEV features (\(P_B\)), we use a fixed 3D sinusoidal positional embedding, following~\cite{wang2021end}.
For the map tokens (\(P_M\)), we use learnable positional embeddings used in frame-level MapNet~\cite{liao2023maptrv2} with newly defined learnable temporal positional embeddings.
For the clip tokens (\(P_C\)), we define new learnable positional embeddings.
Similarly, learnable positional embeddings are defined for the memory tokens that are used in read and write of the Inter-clip Unveiler.

\subsubsection{Loss}
Since our model is built on top of the frame-level MapNet (MapTRv2~\cite{liao2023maptrv2}), we basically follow the loss functions used in MapTR~\cite{liao2023maptr} and MapTRv2~\cite{liao2023maptrv2}. Specifically, we employ the overall loss functions as
\begin{equation}
    \mathcal L_{MapUnveiler} = \lambda_c^M\mathcal L_{cls}^M + \lambda_p^M\mathcal L_{p2p}^M + \lambda_d^M\mathcal L_{dir}^M + \lambda_s^M\mathcal L^M_{PVSeg},
\end{equation}
\begin{equation}
    \mathcal L_{Frame\_MapNet} = \mathcal L_{one2one} + \mathcal L_{one2many} +  \mathcal L_{dense},
\end{equation}
\begin{equation}
    \mathcal L_{one2one} = \lambda_c^F\mathcal L_{cls}^F + \lambda_p^F\mathcal L_{p2p}^F + \lambda_d^F\mathcal L_{dir}^F,
\end{equation}
\begin{equation}
    \mathcal L_{dense} = \lambda_t^F\mathcal L_{depth} + \lambda_b^F\mathcal L_{BEVSeg} + \lambda_s^F\mathcal L_{PVSeg},
\end{equation}
where $\mathcal L_{MapUnveiler}$ and $\mathcal L_{Frame\_MapNet}$ indicate the loss functions for training MapUnveiler and frame-level mapnet, respectively.
$\mathcal L_{cls}$, $\mathcal L_{p2p}$, $\mathcal L_{dir}$, and $\mathcal L_{PVSeg}$ denote classification loss~\cite{liao2023maptr}, point-to-point loss~\cite{liao2023maptr}, edge direction loss~\cite{liao2023maptr}, and PV segmentation loss~\cite{liao2023maptrv2}, respectively. $\mathcal L_{one2one}$, $\mathcal L_{one2many}$, and $\mathcal L_{dense}$ are used for training frame-level MapNet and denote one-to-one loss~\cite{liao2023maptr}, one-to-many loss~\cite{liao2023maptrv2}, and auxiliary dense prediction loss~\cite{liao2023maptrv2}, respectively. $L_{depth}$ and $\mathcal L_{BEVSeg}$ denote depth prediction loss~\cite{liao2023maptrv2} and BEV segmentation loss~\cite{liao2023maptrv2}, respectively. We set hyperparameters to $\lambda_c^M=2$, $\lambda_p^M=5$, $\lambda_d^M=0.005$, $\lambda_s^M=2$, $\lambda_c^F=2$, $\lambda_p^F=5$, $\lambda_d^F=0.005$, $\lambda_t^F=3$, $\lambda_b^F=1$, and $\lambda_s^F=2$.
Note that we did not use one2many loss~\cite{liao2023maptrv2} to save GPU memory during the main training.

\section{Experiments}
\label{section4}
\subsection{Dataset and Metric}
\label{subsection4:Dataset}
We construct experiments on two standard VHC benchmarks: nuScenes~\cite{caesar2020nuscenes} and Argoverse2~\cite{Argoverse2} datasets.
nuScenes offers synchronized six-view images with high-quality HD maps.
It provides 1,000 scenes and each scene consists of about 20 seconds.
In this dataset, we follow the official scene split of 700, 150, and 150 for training, validation, and testing, respectively.
Similarly Argoverse2 provides synchronized seven-view images with high-quality HD maps.
It contains 1,000 scenarios and each scenario consists of about 15 seconds.
Argoverse2 dataset provides the official scene split of 700, 150, and 150 for training, validation, and testing, respectively, which we follow.

For each dataset, we construct maps in two perception ranges: a standard range of 60$\times$30$m$ and a longer range of  100$\times$50$m$.
Additionally, we create challenging validation splits to demonstrate the efficacy of our model under heavy occlusions.
Specifically, we collect the occluded frames if any dynamic objects exists within 2.5$m$ around the ego vehicle.
Thankfully, nuScenes provides 3D cuboid annotations for vehicles and pedestrians, allowing us to automatically create the new challenging split, and the result with this split is given in Sec.~\ref{subsection4:Analysis}.

For fair comparisons with state-of-the-art VHC methods, we follow the standard metric~\cite{li2022hdmapnet,liu2023vectormapnet} of average precision (AP) under several Chamfer thresholds $\{0.5m, 1.0m, 1.5m\}$.
We report the average AP in three Chamfer thresholds for each semantic map categories of pedestrian crossing, divider, and boundary.
We average AP at three Chamfer thresholds and report the results for each semantic map category: pedestrian crossing, divider, and boundary.
To validate the scalability in map categories, we further examine our model by learning semantic centerline, and the results are given in Appendix (see Sec.~\ref{subsubsectionA:extension_centerline}).

\subsection{Implementation Details}
\label{subsection4:Implementation_detail}
Our model is trained with 8 NVIDIA A100 GPUs using batch size of 16.
We train \framework{} using AdamW~\cite{loshchilov2019decoupled} with a learning rate of \(6 \times 10^{-4}\) and a weight decay of 0.01.
We follow the standard setup~\cite{liao2023maptr,liao2023maptrv2} that training models for the total epochs of 24 with an image resolution of \(800 \times 450\) and the total epochs of 6 with an image resolution of \(614 \times 614\) on nuScenes and Argoverse2 datasets, respectively.
The number of the embedding dimension $C$, memory tokens $M$, and clip tokens \(N_{c}\), map elements $N_{i}$, points per map element $N_{p}$ are set to 256, 96, 50, 50, and 20 respectively.
We use temporal window size $T$ and stride $S$ of 3 and 2, respectively.
The spatial size of the BEV feature is \(100 \times 200\).
We pre-train the frame-level MapNet on the standard setting and then main train our model to facilitate the exploitation of the frame-level map information.
For fair comparisons, MapUnveiler is pre-trained for 12 and 3 epochs, and then main trained for an additional 12 and 3 epochs, resulting in a total of 24 and 6 epochs on nuScenes and Argoverse2 datasets, respectively.
During main training, we perform clip-level inference three times and compute losses after each clip-level inference to effectively handle the GPU memory overhead.

\begin{table}[]
\setlength{\tabcolsep}{3pt}
\centering
\caption{Comparisons on nuScenes and Argoverse2 \texttt{val} sets.
AP$_{p}$, AP$_{d}$, AP$_{b}$ indicate the average precision for pedestrian crossing, divider, and boundary, respectively.
FPS is measured using a single NVIDIA A100 GPU.
* are taken from the corresponding papers and are scaled based on the FPS of MapTRv2~\cite{liao2023maptrv2} for a fair comparison.
}
\label{tab:comparison_nuscenes}
\resizebox{\linewidth}{!}{
\begin{tabular}{c|l|cccccc|ccccc}
\toprule
\multirow{2}{*}{Range}                  & \multirow{2}{*}{Method}                                                          & \multicolumn{6}{c|}{nuScenes}                                                     & \multicolumn{5}{c}{Argoverse2}                                                             \\
                                        &                                                                                  & Epoch & AP$_{p}$       & AP$_{d}$       & AP$_{b}$       & mAP            & FPS   & \multicolumn{1}{c}{Epoch} & AP$_{p}$      & AP$_{d}$      & AP$_{b}$      & mAP            \\ 
                                        \midrule
\multirow{12}{*}{\rb{60 $\times$ 30 $m$}} & MapTR{\scriptsize\textcolor{gray}{[ICLR'23]}}~\cite{liao2023maptr}               & 24    & 46.3           & 51.5           & 53.1           & 50.3           & 16.7* & 6                         & 54.7          & 58.1          & 56.7          & 56.5           \\
                                        & MapVR{\scriptsize\textcolor{gray}{[NeurIPS'23]}}~\cite{zhang2023online}          & 24      & 47.7           & 54.4           & 51.4           & 51.2           & 16.7* & -                         & 54.6          & 60.0          & 58.0          & 57.5           \\
                                        & GeMap{\scriptsize\textcolor{gray}{[ECCV'24]}}~\cite{zhang2023online2}            & 24      & 49.2           & 53.6           & 54.8           & 52.6           & 13.2* & -                         & -             & -             & -             & -              \\
                                        & PivotNet{\scriptsize\textcolor{gray}{[ICCV'23]}}~\cite{ding2023pivotnet}         & 24    & 56.2           & 56.5           & 60.1           & 57.6           & 11.1* & -                         & -             & -             & -             & -              \\
                                        & BeMapNet{\scriptsize\textcolor{gray}{[CVPR'23]}}~\cite{qiao2023end}              & 30    & 57.7 & 62.3& 59.4& 59.8& 9.7*  & -                         & -             & -             & -             & -              \\
                                        & MapTRv2{\scriptsize\textcolor{gray}{[IJCV'24]}}~\cite{liao2023maptrv2}          & 24      & 59.8           & 62.4           & 62.4           & 61.5           & 15.6  & 6                         & 62.9          & 72.1          & 67.1          & 67.4           \\
                                        & StreamMapNet{\scriptsize\textcolor{gray}{[WACV'24]}}~\cite{yuan2024streammapnet} & 24    & -              & -              & -              & 62.9           & 12.5* & 6                        & 62.0    & 59.5    & 63.0    & 61.5     \\
                                        & SQD-MapNet{\scriptsize\textcolor{gray}{[preprint]}}~\cite{wang2024stream}        & 24    & 63.0           & 62.5           & 63.3           & 63.9           & -     & 6                        & 64.9    & 60.2    & 64.9    & 63.3     \\
                                        & MGMap{\scriptsize\textcolor{gray}{[CVPR'24]}}~\cite{liu2024mgmap}               & 24      & 61.8           & 65.0           & 67.5           & 64.8           & 12.3* & -                         & -             & -             & -             & -              \\
                                        & MapQR{\scriptsize\textcolor{gray}{[ECCV'24]}}~\cite{liu2024leveraging}          & 24      & 63.4           & 68.0           & 67.7           & 66.4           & 14.2* & 6                           & 64.3          & 72.3          & 68.1          & 68.2           \\
                                        & HIMap{\scriptsize\textcolor{gray}{[CVPR'24]}}~\cite{zhou2024himap}              & 30    & 62.6 & \textbf{68.4} & \textbf{69.1} & 66.7 & 9.7*  & 6                         & \textbf{69.0} & 69.5          & \textbf{70.3} & 69.6           \\
                                        & \framework{} (ours)                                                              & 24    & \textbf{67.6}  & 67.6           & 68.8           & \textbf{68.0}  & 12.7  & 6                         & 68.9          & \textbf{73.7} & 68.9          & \textbf{70.5}  \\
                                        \midrule
\multirow{5}{*}{\rb{100 $\times$ 50 $m$}} & MapTR{\scriptsize\textcolor{gray}{[ICLR'23]}}~\cite{liao2023maptr}               & 24    & 45.5           & 47.1           & 43.9           & 45.5           & 16.7* & 6                         & -             & -             & -             & 47.5 \\
                                        & MapTRv2{\scriptsize\textcolor{gray}{[IJCV'24]}}~\cite{liao2023maptrv2}          & 24    & 58.1           & 61.0           & 56.6           & 58.6           & 15.6  & 6                         & 66.2          & 61.4          & 54.1          & 60.6           \\
                                        & StreamMapNet{\scriptsize\textcolor{gray}{[WACV'24]}}~\cite{yuan2024streammapnet} & 24    & 62.9           & 63.1           & 55.8           & 60.6           & 12.5* & 6                        & -             & -             & -             & 57.7     \\
                                        & SQD-MapNet{\scriptsize\textcolor{gray}{[preprint]}}~\cite{wang2024stream}        & 24    & 67.0           & 65.5           & 59.5           & 64.0           & -     & 6                        & 66.9    & 54.9    & 56.1    & 59.3     \\
                                        & \framework{} (ours)                                                              & 24    & \textbf{68.0}  & \textbf{70.0}  & \textbf{68.2}  & \textbf{68.7}  & 12.7  & 6                         & \textbf{69.7} & \textbf{67.1} & \textbf{59.3} & \textbf{65.4}  \\
                                        \bottomrule
\end{tabular}
}
\vspace{-2mm}
\end{table}

\subsection{Comparisons}
\label{subsection4:Comparisons}
We compare our \framework{} against state-of-the-art VHC methods trained with standard settings, which use ResNet50~\cite{he2016deep} as a backbone, multi-camera modality, and train for 24 epochs and 6 epochs on nuScenes~\cite{caesar2020nuscenes} and Argoverse2~\cite{Argoverse2} respectively.
As shown in Tab.~\ref{tab:comparison_nuscenes}, \framework{} achieves state-of-the-art performance on all validation sets.
Specifically, we surpass the state-of-the-art temporal model (SQD-MapNet~\cite{wang2024stream}) by mAP of 4.1\% and 4.7\% on two range settings of nuScenes validation set and 7.2\% and 6.1\% on Argoverse2.
We also outperform a heavy VHC model (HiMap~\cite{zhou2024himap}) by 1.3\% and 0.9\% on two benchmark sets.
Notably, we boost frame-level model (MapTRv2~\cite{liao2023maptrv2}) by 6.5\% and 10.1\% on nuScenes and 3.1\% and 4.8\% on Argoverse2.
The superiority of our approach is particularly evident in the long-range setting: mAP on 60$\times$30$m$$\rightarrow$100$\times$50$m$ settings are 61.5\%$\rightarrow$58.6\% (-2.9\%) and 68.0\%$\rightarrow$68.7\% (+0.7\%) for MapTRv2~\cite{liao2023maptrv2} and ours.
Although \framework{} incorporates temporal modules, we achieve a reasonable inference speed (12.7 FPS) compared to frame-level MapNet (MapTRv2~\cite{liao2023maptrv2}, 15.6 FPS),
surpassing both performance and speed compared to the state-of-the-art (HiMap~\cite{zhou2024himap}, 9.7 FPS).

\begin{figure}
    \centering
    \includegraphics[width=1\linewidth]{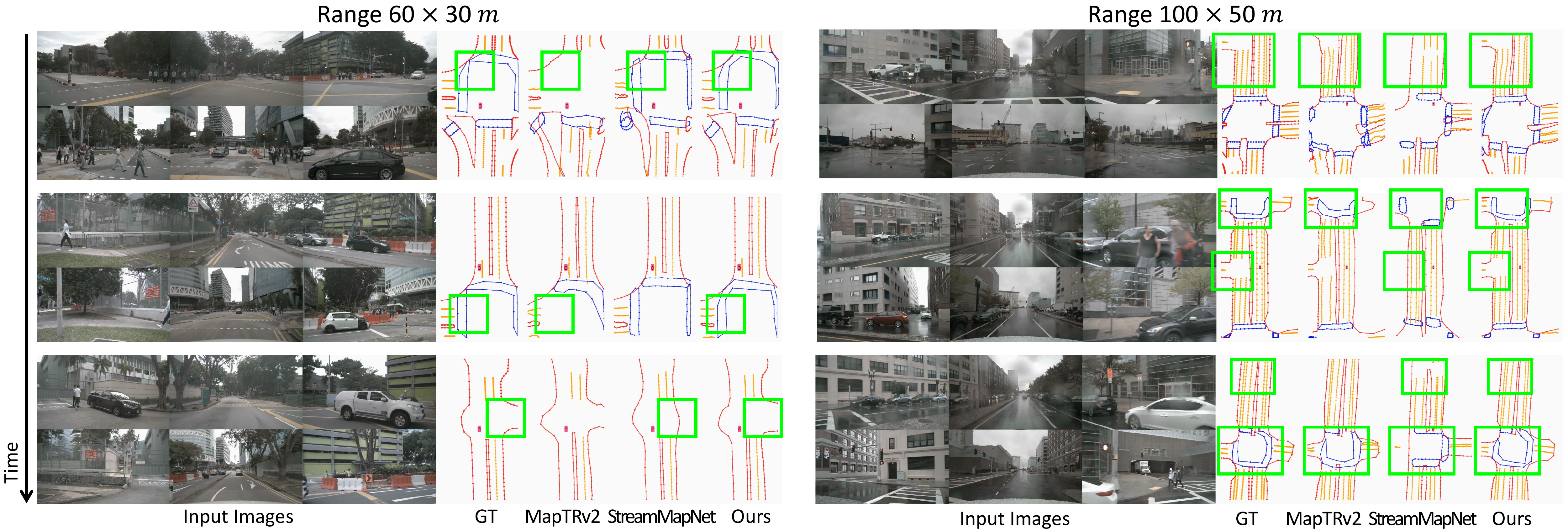}
    \caption{
    Qualitative comparisons on two range variants of nuScenes \texttt{val} set: 60$\times$30$m$ and 100$\times$50$m$.
    We compare our \framework{} with MapTRv2~\cite{liao2023maptrv2} and StreamMapNet~\cite{yuan2024streammapnet}.
    We marked significant improvements from MapTRv2 and StreamMapNet using green boxes.
    }
    \label{fig:qualitative_comparison}
\end{figure}

\vspace{-2mm}
\paragraph{Qualitative Comparison.}
Fig.~\ref{fig:qualitative_comparison} shows qualitative comparisons on two perception range settings on nuScenes benchmark.
We compare our \framework{} with the state-of-the-art frame-level model (MapTRv2~\cite{liao2023maptrv2}) and temporal model (StreamMapNet~\cite{yuan2024streammapnet}).
As shown in the figure, MapTRv2 often mis-detects complex pedestrian crossings and cannot precisely predict dividers occluded by vehicles on the road.
StreamMapNet struggles to accurately predict the boundaries of intersections, despite leveraging temporal information.
In contrast, \framework{} consistently delivers accurate results for all map categories in most cases.
We provide more qualitative results on various scenes in Appendix and full-frame results through the supplementary video.

\vspace{-2mm}
\subsection{Analysis Experiments}
\label{subsection4:Analysis}
\vspace{-2mm}
In this study, we present extensive experimental results on nuScenes benchmark and provide analysis.

\begin{table}
\setlength{\tabcolsep}{5pt}
\centering
\vspace{-2mm}
\caption{Experimental results of the long training schedules.}
\label{tab:long_schedule}
\begin{tabular}{c|ccccc|ccccc}
\toprule
\multirow{2}{*}{Range}     & \multicolumn{5}{c|}{nuScenes}               & \multicolumn{5}{c}{Argoverse2}       \\
                           & Epoch & AP$_{p}$ & AP$_{d}$ & AP$_{b}$ & mAP    & Epoch & AP$_{p}$ & AP$_{d}$ & AP$_{b}$ & mAP  \\ \midrule
\multirow{3}{*}{60 $\times$ 30 $m$}  & 24    & 67.6  & 67.6  & 68.8  & 68.0  & 6     & 68.9  & 73.7  & 68.9  & 70.5 \\
                           & 48    & 69.5  & 69.4  & 70.5  & 69.8 & 12     & 69.0  & 74.9  & 69.1  & 71.0 \\
                           & 110   & 71.0    & 69.1  & 71.8  & 70.6  & 30    & 72.5  & 74.2  & 71.9  & 72.9 \\ \midrule
\multirow{3}{*}{100 $\times$ 50 $m$} & 24    & 68.0    & 70.0    & 68.2  & 68.7  & 6     & 69.7  & 67.1  & 59.3  & 65.4 \\
                           & 48    & 68.4  & 71.2  & 68.3  & 69.3  & 12     & 70.4  & 66.8  & 59.3  & 65.5 \\
                           & 110   & 71.2  & 71.7  & 72.2  & 71.7  & 30    & 71.7  & 67.9  & 62.6  & 67.4 \\ \bottomrule
\end{tabular}
\vspace{-2mm}
\end{table}

\vspace{-2mm}
\paragraph{Long Training Schedules.}
Tab.~\ref{tab:long_schedule} demonstrates the performance improvement achieved through longer training schedules.
We obtained (24, 48, 110) and (6, 12, 30) epoch models by pre-training for (12, 24, 24) and (3, 6, 6) epochs, and then main training for (12, 24, 86) and (3, 6, 24) epochs on nuScenes and Argoverse2 datasets, respectively.
As given in the table, the performance consistently improves with longer training, although the gains are relatively marginal.
We conjecture that our MapUnveiler converges rapidly because we train the temporal modules from a pre-trained frame-level MapNet.
Since our model shows no significant gain from the 110 epoch model on nuScenes, we opt for 48 epochs for the remaining analysis experiments.

\begin{table}
	\begin{minipage}[t]{0.53\linewidth}
\setlength{\tabcolsep}{5pt}
\centering
\caption{Experimental results under heavy occlusions.
}
\label{tab:occlusion}
\begin{tabular}{l|cccc}
\toprule
Method                                   & AP$_{p}$      & AP$_{d}$      & AP$_{b}$      & mAP           \\
\midrule
MapTRv2~\cite{liao2023maptrv2}           & 24.1          & 62.2          & 55.2          & 47.1          \\
StreamMapNet~\cite{yuan2024streammapnet} & 32.1          & 59.4          & 67.9          & 53.1          \\
\framework{} (ours)                      & \textbf{38.6} & \textbf{80.6} & \textbf{72.2} & \textbf{63.8}\\
\bottomrule
\end{tabular}
\setlength{\tabcolsep}{4.7pt}
\centering
\vspace{1mm}
\caption{Ablation study on the proposed modules.
}
\label{tab:module_ablation}
\begin{tabular}{l|cccc}
\toprule
Method                         & AP$_{p}$      & AP$_{d}$      & AP$_{b}$      & mAP             \\
\midrule
MapTRv2~\cite{liao2023maptrv2} & 58.8          & 61.8          & 62.8          & 61.2         \\
+ Intra-clip Unveiler          & 65.6          & 67.6          & 68.0          & 67.1        \\
+ Inter-clip Unveiler          & \textbf{69.5} & \textbf{69.4} & \textbf{70.5} & \textbf{69.8} \\
\bottomrule
\end{tabular}
\setlength{\tabcolsep}{2.4pt}
\centering
\vspace{1mm}
\caption{Speed analysis of the proposed modules.
}
\label{tab:module_ablation_fps}
\begin{tabular}{l|ccccccc}
\toprule
Method                        & FPS & GPU{\scriptsize~(MB)} & Params{\scriptsize~(MB)}  \\
\midrule
MapTRv2~\cite{liao2023maptrv2}    & 15.6 & 830.4 & 76.4 \\
+ Intra-clip Unveiler          & 13.1 & 1552.5 & 144.0 \\
+ Inter-clip Unveiler          & 12.7 & 1614.9 & 213.9 \\
\bottomrule
\end{tabular}
\setlength{\tabcolsep}{8.3pt}
\centering
\vspace{1mm}
\caption{Results with freezing frame-level MapNet.
}
\label{tab:freeze}
\begin{tabular}{l|cccc}
\toprule
Read                 & AP$_{p}$      & AP$_{d}$      & AP$_{b}$      & mAP           \\
\midrule
Freeze               & 64.9          & 66.5          & 68.7          & 66.7          \\
End-to-End & \textbf{69.5} & \textbf{69.4} & \textbf{70.5} & \textbf{69.8} \\
\bottomrule
\end{tabular}
\setlength{\tabcolsep}{6.3pt}
\centering
\vspace{1mm}
\caption{Input variants in read.
}
\label{tab:ablation_read}
\begin{tabular}{l|cccc}
\toprule
Read                 & AP$_{p}$      & AP$_{d}$      & AP$_{b}$      & mAP           \\
\midrule
Memory               & 68.3          & 68.5          & 70.0          & 68.9          \\
Memory + $Q^{map}$ & \textbf{69.5} & \textbf{69.4} & \textbf{70.5} & \textbf{69.8} \\
\bottomrule
\end{tabular}
    \end{minipage}
    \hfill
    \begin{minipage}[t]{0.47\linewidth}
\setlength{\tabcolsep}{1.2pt}
\centering
\caption{Input variants in write.
}
\label{tab:ablation_write}
\begin{tabular}{l|cccc}
\toprule
Write                          & AP$_{p}$      & AP$_{d}$      & AP$_{b}$      & mAP           \\
\midrule
None                           & 65.6          & 67.6          & 68.0          & 67.1          \\
$U_{L}^{clip}$                    & 68.7          & 68.9          & 69.6          & 69.1          \\
$U_{L}^{clip}$ + $U_{L}^{map}$       & \textbf{69.5} & \textbf{69.4} & \textbf{70.5} & \textbf{69.8} \\
$U_{L}^{clip}$ + $U_{L}^{map}$ + $U_{L}^{BEV}$ & 68.5          & 68.8          & 69.5          & 68.9          \\
\bottomrule
\end{tabular}
\setlength{\tabcolsep}{7.6pt}
\centering
\vspace{-1mm}
\caption{Temporal window size $T$ and stride $S$.
}
\label{tab:ablation_temporal_window_stride}
\begin{tabular}{cc|cccc}
\toprule
$T$ & $S$ & AP$_{p}$ & AP$_{d}$ & AP$_{b}$ & mAP  \\
\midrule
1   & 1   & 63.5     & 66.0     & 66.1     & 65.2 \\
3   & 1   & 69.0     & 69.4     & 69.4     & 69.3 \\
3   & 2   & 69.5     & 69.4     & \textbf{70.5}     & 69.8 \\
3   & 3   & 68.6     & 68.7     & 69.6     & 68.9 \\
5   & 3   & \textbf{70.4}     & \textbf{69.5}     & \textbf{70.5}     & \textbf{70.1} \\
\bottomrule
\end{tabular}
\setlength{\tabcolsep}{9pt}
\centering
\vspace{-1mm}
\caption{Variant memory token sizes.
}
\label{tab:ablation_memory_token_size}
\begin{tabular}{c|cccc}
\toprule
$M$ & AP$_{p}$ & AP$_{d}$ & AP$_{b}$ & mAP  \\
\midrule
24  & 68.2     & 68.3     & 69.1     & 68.5 \\
48  & 68.4     & 69.3     & 69.9     & 69.2 \\
96  & 69.5     & \textbf{69.4}     & \textbf{70.5}     & \textbf{69.8} \\
192 & 69.0     & \textbf{69.4}     & 70.1     & 69.5 \\
384 & \textbf{70.0}     & 69.3     & 69.9     & 69.7 \\
\bottomrule
\end{tabular}
\setlength{\tabcolsep}{9pt}
\centering
\vspace{-1mm}
\caption{Variant clip token sizes
}
\label{tab:ablation_clip_token_size}
\begin{tabular}{c|cccc}
\toprule
$N_{c}$ & AP$_{p}$ & AP$_{d}$ & AP$_{b}$ & mAP  \\
\midrule
25      & 67.8     & 69.6     & 70.1     & 69.2 \\
50      & \textbf{69.5}     & \textbf{69.4}     & \textbf{70.5}     & \textbf{69.8} \\
100     & 67.9     & 68.1     & 69.6     & 68.5 \\
200     & 69.3     & 68.8     & 70.2     & 69.4 \\
\bottomrule
\end{tabular}
    \end{minipage}
    \hfill
\vspace{-2mm}
\end{table}

\vspace{-2mm}
\paragraph{Robustness to Occlusion.}
Tab.~\ref{tab:occlusion} presents results on our challenging validation splits collected based on the nearest dynamic objects (detailed in Sec.~\ref{subsection4:Dataset}).
For comparisons, we evaluate MapTRv2 and StreamMapNet using the code and pre-trained weight provided in each official repository.
As shown in the table, \framework{} surpasses MapTRv2 and StreamMapNet in all evaluated metrics.
In particular, existing approaches degrade performance significantly compared to the results on the standard split, \ie, 61.5\%$\rightarrow$47.1\% (-14.4\%) in MapTRv2 and 62.9\%$\rightarrow$53.1\% (-9.8\%) in StreamMapNet, 
\framework{} also shows a performance degradation of 69.8\%→63.8\% (-6.0\%), but it demonstrates a smaller performance gap compared to previous studies.

\vspace{-2mm}
\paragraph{Effectiveness of The Proposed Modules.}
Tabs.~\ref{tab:module_ablation} and~\ref{tab:module_ablation_fps} show an ablation study and speed analysis of our two proposed modules: Intra-clip Unveiler and Inter-clip Unveiler.
We started with a frame-level MapNet, MapTRv2~\cite{liao2023maptrv2}, obtaining the result by reproducing it on our system.
As shown in the result, Intra-clip Unveiler significantly boosts the performance from 61.2\% to 67.1\% (+5.9\%) by mapping within a clip, and it rarely requires extra computation (15.6 FPS $\rightarrow$ 13.1 FPS).
Inter-clip Unveiler further improves performance from 67.1\% to 69.8\% (+2.7\%) by building global relationship of the mapping.
However, our approach requires approximately two times more peak GPU memory and three times more parameters compared to the frame-level MapTRv2 model during inference. This could be considered a potential limitation of our method, but the amounts are not large.

\vspace{-2mm}
\paragraph{Frozen Frame-level MapNet.}
Tab.~\ref{tab:freeze} presents the results of \framework{} where frame-level MapNet is frozen and only Intra-clip Unveiler and Inter-clip Unveiler are trained.
Despite the frozen model consuming the same amount of computation during inference, it effectively reduces GPU memory consumption and converges quickly during training.
As given in the table, we obtained slightly lower performance compared to the end-to-end training setting.
Interestingly, however, the frozen model also achieves the state-of-the-art performance on nuScenes benchmark (current SOTA performance is mAP of 66.7\%, the concurrent work of HIMap~\cite{zhou2024himap}).
This indicates that our approach effectively unveils the hidden maps in the frame-level representations and does not rely solely on strong BEV representation learning.

\vspace{-2mm}
\paragraph{Ablation Study on Read and Write.}
As given in Tab.~\ref{tab:ablation_read}, reading only from memory is not enough, and feeding map tokens facilitates the understanding of the current map information (68.9\%$\rightarrow$69.8\%).
We also ablate the input in write, and the result is given in Tab.~\ref{tab:ablation_write}.
We significantly improve the performance by writing clip tokens to the memory (67.1\%$\rightarrow$69.1\%).
This suggests that the propagation of memory tokens facilitates constructing the global map.
By additionally writing map tokens, we further boost the performance by explicitly providing the current map information (69.1\%$\rightarrow$69.8\%).
Unfortunately, however, we cannot obtain performance gain by writing dense BEV feature additionally.
We conjecture that memorizing different types of features, \ie, vectorized representation of tokens and rasterized representation of dense features, may disturb training and tend to converge to a sub-optimal state.

\vspace{-2mm}
\paragraph{Temporal Window Size $T$ and Stride $S$.}
We train \framework{} with various temporal window size $T$ and stride $S$.
As shown in Tab.~\ref{tab:ablation_temporal_window_stride}, increasing the temporal window leads to performance improvement by interacting with more frames in a clip.
However, training with a $T=5$ setting requires $>$ 40GB of GPU memory, limiting the GPU models.
If we train \framework{} with $T=3$, it consumes $<$ 32GB of GPU memory, making it possible to train with various GPU models, and it achieves comparable performance.
Additionally, selecting either too short ($S=1$) or too long ($S=3$) a temporal stride yields sub-optimal results.
Therefore, we opt for $T=3$ and $S=2$ as the default setting.

\vspace{-2mm}
\paragraph{Memory Token Size $M$.}
Tab.~\ref{tab:ablation_memory_token_size} presents the experimental results with various memory token sizes.
Selecting a memory token size of 96 or above does not significantly change the performance, whereas choosing a size smaller than 96 results in considerable performance degradation.
This indicates that \framework{} does not require a large memory token size to store the 50 clip tokens and 50$\times$20 map tokens, demonstrating that it performs memory-efficiently.

\vspace{-2mm}
\paragraph{Clip Token Size $N_{c}$.}
Tab.~\ref{tab:ablation_clip_token_size} gives the results with various clip token sizes.
\framework{} achieves state-of-the-art performance of 69.2\% using only 25 clip tokens.
We further boost the performance by using a larger clip token size of 50.
However, employing clip token sizes beyond 50 disrupts learning and leads to a performance degradation.

\begin{wrapfigure}{r}{.5\linewidth}
    \centering
    \vspace{-10mm}
    \includegraphics[width=\linewidth]{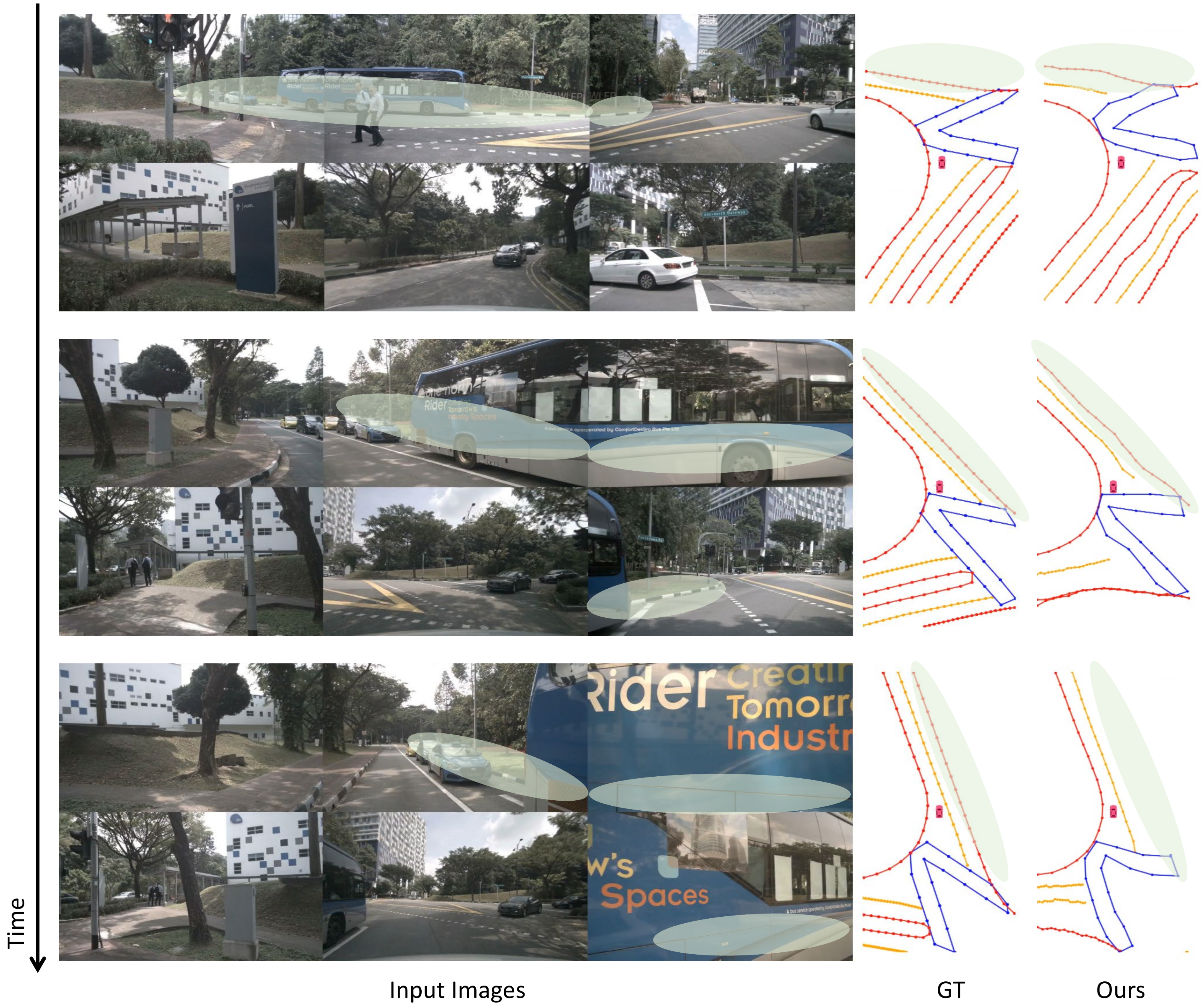}
    \caption{
    A limitation under heavy occlusions.
    MapUnveiler initially roughly predicts the boundary where we highlighted in green. However, the invisible regions are continuous, and MapUnveiler eventually predicts the region as having no boundary.
    }
    \vspace{-17mm}
    \label{fig:fig_failure_case_1}
\end{wrapfigure}

\vspace{-2mm}
\paragraph{More Analysis in Appendix.}
We provide additional experimental results including centerline, 3D map construction, geo-disjoint split, various backbones, and limitation analysis in Appendix.

\vspace{-2mm}
\subsection{Limitation}
\label{limitations}
\vspace{-2mm}
MapUnveiler takes temporally consecutive frames as input.
Therefore, if an intermediate frame is not correctly inputted due to communication errors in real-world scenarios, the unveiling within the clip may not be performed properly.
We expect that the performance will recover quickly from the subsequent unveiling pipeline.
Additionally, if the model cannot see a clear region across all frames as shown in Fig.~\ref{fig:fig_failure_case_1}, MapUnveiler may fail to recognize the occluded map information.

\section{Conclusion}
\label{section5}
\vspace{-2mm}
In this paper, we present a new VHC paradigm that constructs clip-level maps to efficiently incorporate conventional offline mapping strategies.
In contrast to the recent temporal VHC models that do not consider the cumulative impacts of occluded map elements and directly propagate the noisy BEV features, we unveil the hidden map and noise in BEV features by interacting with compact clip tokens.
To establish global mapping efficiently, we propagate the clip tokens instead of dense BEV features.
With these two advanced modules, we propose \framework{}, which significantly outperforms previous works in challenging scenarios such as long-range perception and heavy occlusions.
Since we introduce a novel insight into online VHC approaches to incorporate mapping strategy efficiently, we hope that our research motivates follow-up studies to delve deeper into establishing global mapping online and leads to practical VHC solutions.

{\small
\bibliographystyle{ieee_fullname}
\bibliography{ref}
}

\newpage
\appendix

\section{Appendix}

\subsection{Clip-level Inference Scheme}
\label{subsubsectionA:definition_inference_scheme}
Our MapUnveiler runs in a clip-level. Specifically, the $k$-th clip of $C_k$ includes
\begin{equation}
C_k =  \lbrace f_t \rbrace_{t=kS+1}^{kS+T}.
\end{equation}
MapUnveiler infers from $k=0$. $f_t$ represents a frame at time $t$, and $\lbrace f_t \rbrace_{t=kS+1}^{kS+T}$ denote a sequence of consecutive frames from $kS+1$ to $kS+T$, where $S$ and $T$ indicate the clip stride and the temporal window, respectively. Therefore, when $T=3$ and $S=2$, we obtain clip sets such as
$C_0=\lbrace f_1, f_2, f_3 \rbrace$, $C_1=\lbrace f_3, f_4, f_5 \rbrace$, ..., and  $C_k=\lbrace f_{2k+1}, f_{2k+2}, f_{2k+3}\rbrace$.

\subsection{Predicting with Centerline}
\label{subsubsectionA:extension_centerline}
In Tab.~\ref{tab:extension_centerline}, we present a quantitative comparison on centerline predictions.
\framework{} outperforms MapTRv2~\cite{liao2023maptrv2} by significant margins on both nuScenes (54.0\%$\rightarrow$63.0\%, +9.0\%) and Argoverse (62.6\%$\rightarrow$68.0\%, +5.4\%) benchmarks.
This demonstrates the robustness of our approach on various map categories.

\begin{table}[h]
\centering
\vspace{-1mm}
\caption{Centerline predictions.
AP$_{p}$, AP$_{d}$, AP$_{b}$, AP$_{c}$ indicate the average precision for pedestrian crossing, divider, boundary, and centerline, respectively.
}
\label{tab:extension_centerline}
\setlength{\tabcolsep}{3pt}
\begin{tabular}{l|cccccc|cccccc}
\toprule
\multirow{2}{*}{Method}        & \multicolumn{6}{c|}{nuScenes}                    & \multicolumn{6}{c}{Argoverse2}                   \\
                               & Epoch & AP$_{p}$ & AP$_{d}$ & AP$_{b}$ & AP$_{c}$ & mAP  & Epoch & AP$_{p}$ & AP$_{d}$ & AP$_{b}$ & AP$_{c}$ & mAP  \\
                               \midrule
MapTRv2{\scriptsize\textcolor{gray}{[IJCV'24]}}~\cite{liao2023maptrv2} & 24 & 50.1     & 53.9     & 58.8     & 53.1     & 54.0 & 6 & 55.2     & 67.2     & 64.8     & 63.2     & 62.6 \\
\framework{} (ours)            & 24 & \textbf{59.0}     & \textbf{65.7}     & \textbf{68.0}     & \textbf{59.4}     & \textbf{63.0} & 6 & \textbf{62.2}     & \textbf{73.2}     & \textbf{70.0}     & \textbf{66.6}     & \textbf{68.0} \\
\bottomrule
\end{tabular}
\end{table}

\subsection{Extension to 3D Vectorized Map Construction}
\label{subsubsectionA:extension_3d}
Tab.~\ref{tab:extension_3d}, presents a comparison on 3D map construction.
In this study, we conduct experiment on Argoverse2 dataset.
\framework{} achieves state-of-the-art performance in Tab.~\ref{tab:extension_3d}, demonstrating the efficacy of our approach even when extended to 3D VHC.

\begin{table}[h]
\centering
\vspace{-1mm}
\caption{Comparisons on Argoverse2 \texttt{val} set with 3D map construction.
}
\label{tab:extension_3d}
\begin{tabular}{l|ccccc}
\toprule
Method                                                                  & Epoch & AP$_{p}$ & AP$_{d}$ & AP$_{b}$ & mAP  \\
\midrule
MapTRv2{\scriptsize\textcolor{gray}{[IJCV'24]}}~\cite{liao2023maptrv2} & 6 & 60.7     & 68.9     & 64.5     & 64.7 \\
HIMap{\scriptsize\textcolor{gray}{[CVPR'24]}}~\cite{zhou2024himap}     & 6 & \textbf{66.7}     & 68.3     & \textbf{70.3}     & 68.4 \\
\framework{}                                                            & 6 & 66.0     & \textbf{72.6}     & 67.6     & \textbf{68.7}\\
\bottomrule
\end{tabular}
\end{table}

\subsection{A More Comparison on Geo-disjoint Split}
\label{subsubsectionA:geo_disjoint}
Tab.~\ref{tab:geo_disjoint} presents a experimental results on a recent geo-disjoint dataset split proposed in StreamMapNet~\cite{yuan2024streammapnet}.
In this dataset, the performance of MapTRv2~\cite{liao2023maptrv2} has been significantly drop from 61.5\% to 36.6\%. 
We also successfully achieve state-of-the-art performance on geo-disjoint dataset split.

\begin{table}[h]
\vspace{-1mm}
\caption{Experimental results on geo-disjoint dataset split proposed in StreamMapNet~\cite{yuan2024streammapnet}. The results are obtained on nuScenes with 60$\times$30$m$ range.}
\centering
\label{tab:geo_disjoint}
\begin{tabular}{l|ccccc}
\toprule
Method                & Epoch & AP$_{p}$ & AP$_{d}$ & AP$_{b}$ & mAP  \\ \midrule
VectorMapNet{\scriptsize\textcolor{gray}{[ICML'23]}}~\cite{liu2023vectormapnet}         & 120   & 15.8     & 17.0     & 21.2     & 18.0 \\
MapTR{\scriptsize\textcolor{gray}{[ICLR'23]}}~\cite{liao2023maptr}                & 24    & 6.4      & 20.7     & 35.5     & 20.9 \\
StreamMapNet{\scriptsize\textcolor{gray}{[WACV'24]}}~\cite{yuan2024streammapnet} & 24    & 29.6     & \textbf{30.1}     & 41.9     & 33.9 \\
MapTRv2{\scriptsize\textcolor{gray}{[IJCV'24]}}~\cite{liao2023maptrv2} & 24    & 37.2     & 26.5     & 46.1     & 36.6 \\
MapUnveiler (ours)  & 24    & \textbf{43.2}       & 26.5       & \textbf{48.7}       & \textbf{39.4}   \\ \bottomrule
\end{tabular}
\end{table}

\subsection{Various Backbones}
\label{subsubsectionA:various_backbones}
We present experimental results with various backbones: ResNet-18~\cite{he2016deep} and V2-99~\cite{lee2020centermask}. As shown in Tab.~\ref{tab:various_backbone}, our method is not limited to MapTRv2~\cite{liao2023maptrv2} with ResNet50, but can be extended to ResNet18 and V2-99 backbones.

\begin{table}[h]
\centering
\setlength{\tabcolsep}{1.5pt}
\caption{Experimental results with various backbones.}
\label{tab:various_backbone}
\begin{tabular}{c|l|c|ccccc|ccccc}
\toprule
\multirow{2}{*}{Range}     & \multirow{2}{*}{Method} & \multirow{2}{*}{Backbone} & \multicolumn{5}{c|}{nuScenes}                                         & \multicolumn{5}{c}{Argoverse2}                                        \\
                           &                         &                           & Epoch & AP$_{p}$      & AP$_{d}$      & AP$_{b}$      & mAP           & Epoch & AP$_{p}$      & AP$_{d}$      & AP$_{b}$      & mAP           \\ \midrule
\multirow{6}{*}{\rb{60 $\times$ 30 $m$}}  & MapTRv2{\scriptsize\textcolor{gray}{[IJCV'24]}}~\cite{liao2023maptrv2}                 & R18                       & 24    & 53.3          & 58.5          & 58.5          & 56.8          & 6     & 58.8          & 68.5          & 64            & 63.8          \\
                           & MapTRv2{\scriptsize\textcolor{gray}{[IJCV'24]}}~\cite{liao2023maptrv2}                 & R50                     & 24    & 59.8          & 62.4          & 62.4          & 61.5          & 6     & 62.9          & 72.1          & 67.1          & 67.4          \\
                           & MapTRv2{\scriptsize\textcolor{gray}{[IJCV'24]}}~\cite{liao2023maptrv2}                 & V2-99                     & 24    & 63.6          & 67.1          & 69.2          & 66.6          & 6     & 64.5          & 72.2          & 70.1          & 68.9          \\
                           & MapUnveiler (ours)      & R18                       & 24    & 62.4          & 65.2          & 65.7          & 64.4          & 6     & 63.8          & 70.1          & 67.1          & 67.0            \\
                           & MapUnveiler (ours)      & R50                       & 24    & 67.6          & 67.6          & 68.8          & 68.0          & 6     & 68.9          & 73.7          & 68.9          & 70.5          \\
                           & MapUnveiler (ours)      & V2-99                     & 24    & \textbf{69.8} & \textbf{72.0}   & \textbf{74.7} & \textbf{72.1} & 6     & \textbf{69.6} & \textbf{75.1} & \textbf{72.8} & \textbf{72.5} \\ \midrule
\multirow{6}{*}{\rb{100 $\times$ 50 $m$}} & MapTRv2{\scriptsize\textcolor{gray}{[IJCV'24]}}~\cite{liao2023maptrv2}                  & R18                       & 24    & 52.7          & 57.3          & 51.5          & 53.8          & 6     & 60.3          & 57.6          & 49.6          & 55.8          \\
                           & MapTRv2{\scriptsize\textcolor{gray}{[IJCV'24]}}~\cite{liao2023maptrv2}                  & R50                     & 24    & 58.1          & 61.0          & 56.6          & 58.6          & 6     & 66.2          & 61.4          & 54.1          & 60.6          \\
                           & MapTRv2{\scriptsize\textcolor{gray}{[IJCV'24]}}~\cite{liao2023maptrv2}                  & V2-99                     & 24    & 62.6          & 67.8          & 65.2          & 65.2          & 6     & 68.5          & 62.1          & 58.4          & 63.0          \\
                           & MapUnveiler (ours)      & R18                       & 24    & 64.1          & 67.3          & 65.3          & 65.6          & 6     & 65.4          & 63.2          & 54.7          & 61.1          \\
                           & MapUnveiler (ours)      & R50                       & 24    & 68.0          & 70.0          & 68.2          & 68.7          & 6     & 69.7          & \textbf{67.1}          & 59.3          & 65.4          \\
                           & MapUnveiler (ours)      & V2-99                     & 24    & \textbf{70.5} & \textbf{74.6} & \textbf{73.6} & \textbf{72.9} & 6     & \textbf{71.1} & 66.1 & \textbf{61.4} & \textbf{66.2} \\ \bottomrule
\end{tabular}
\end{table}

\subsection{Randomly Dropped Intermediate Frames}
\label{subsubsectionA:random_drop}
In the limitations section (Sec.~\ref{limitations}), we discussed that our MapUnveiler relies on temporally consecutive frames, and the performance can degrade when intermediate frames are missing.
However, we expected that the performance would recover quickly from the subsequent unveiling pipeline.
To validate this, we evaluated two models with randomly dropped intermediate frames. Frames were dropped by converting multi-camera images into black images. The experiment was conducted with drop rates of 20\%, 10\%, and 5\%, and the results are given in Tab.~\ref{tab:drop}. MapUnveiler is affected by dropped frames, but the performance degradation is reasonable compared to MapTRv2~\cite{liao2023maptrv2}.

\begin{table}[h]
\centering
\caption{Experimental results with randomly dropped intermediate frames. The results are obtained on nuScenes with 60$\times$30$m$ range
}
\label{tab:drop}
\begin{tabular}{c|c|cccc}
\toprule
Method                              & Drop rate & AP$_{p}$ & AP$_{d}$ & AP$_{b}$ & mAP  \\
\midrule
\multirow{4}{*}{MapTRv2{\scriptsize\textcolor{gray}{[IJCV'24]}}~\cite{liao2023maptrv2}}       & 0\%       & 58.8     & 61.8     & 62.8     & 61.2 \\
                                    & 5\%       & 56.6     & 56.6     & 58.7     & 57.7 \\
                                    & 10\%      & 53.4     & 54.8     & 55.3     & 54.5 \\
                                    & 20\%      & 47.2     & 48.5     & 38.9     & 48.2 \\
                                    \midrule
\multirow{4}{*}{MapUnveiler (Ours)} & 0\%       & 69.5     & 69.4     & 70.5     & 69.8 \\
                                    & 5\%       & 58.0     & 60.6     & 60.0     & 66.9 \\
                                    & 10\%      & 63.3     & 64.7     & 65.0     & 64.3 \\
                                    & 20\%      & 66.2     & 66.9     & 67.6     & 59.6 \\
                                    \bottomrule
\end{tabular}
\end{table}

\subsection{More Qualitative Results}
\label{subsubsectionA:more_qualitative_results}
\begin{figure}
    \centering
    \includegraphics[width=.95\linewidth]{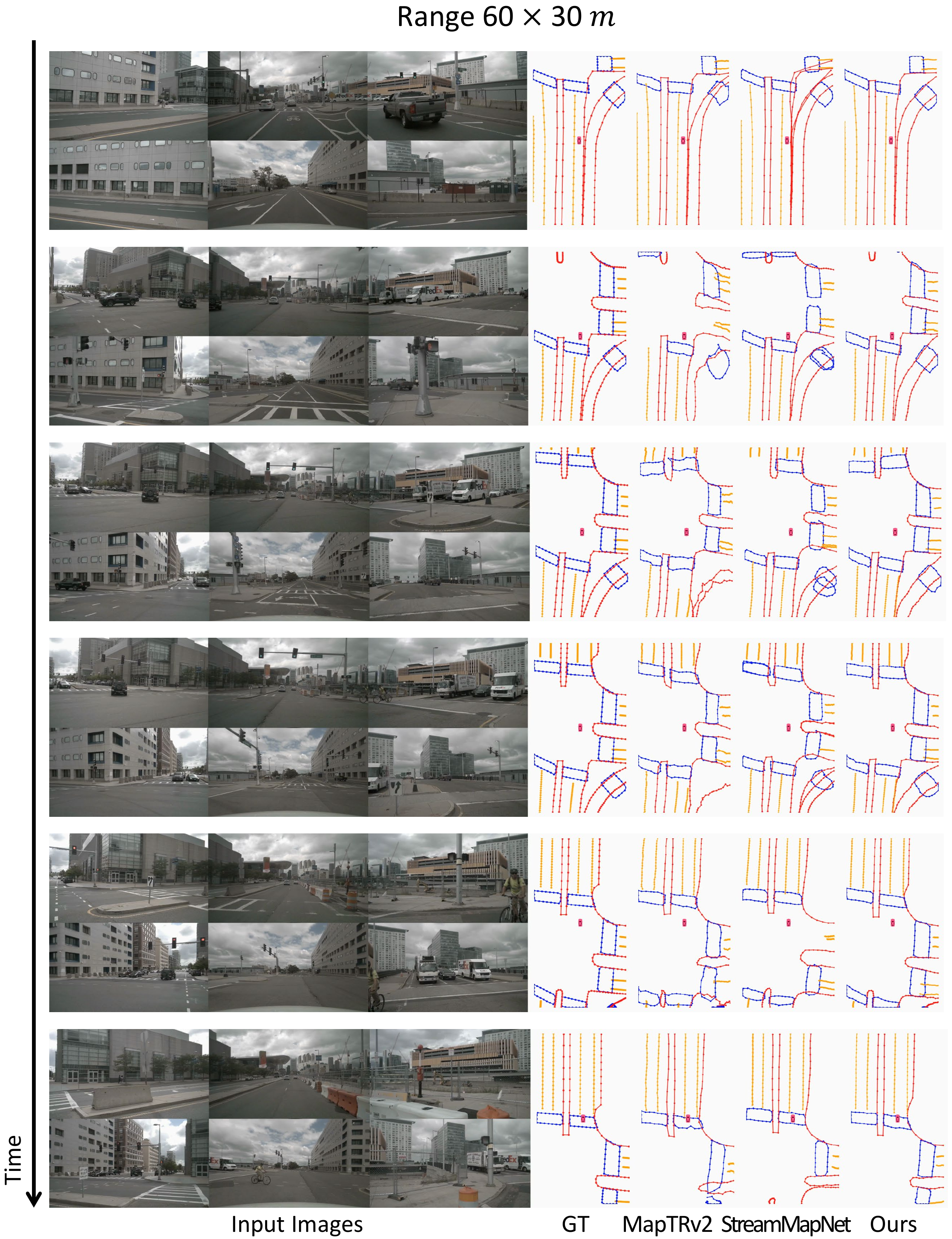}
    \caption{
    Qualitative comparisons on nuScenes \texttt{val} with 60$\times$30$m$ perception range setting.
    }
    \label{fig:fig_qualitative_appendix1}
\end{figure}

\begin{figure}
    \centering
    \includegraphics[width=.95\linewidth]{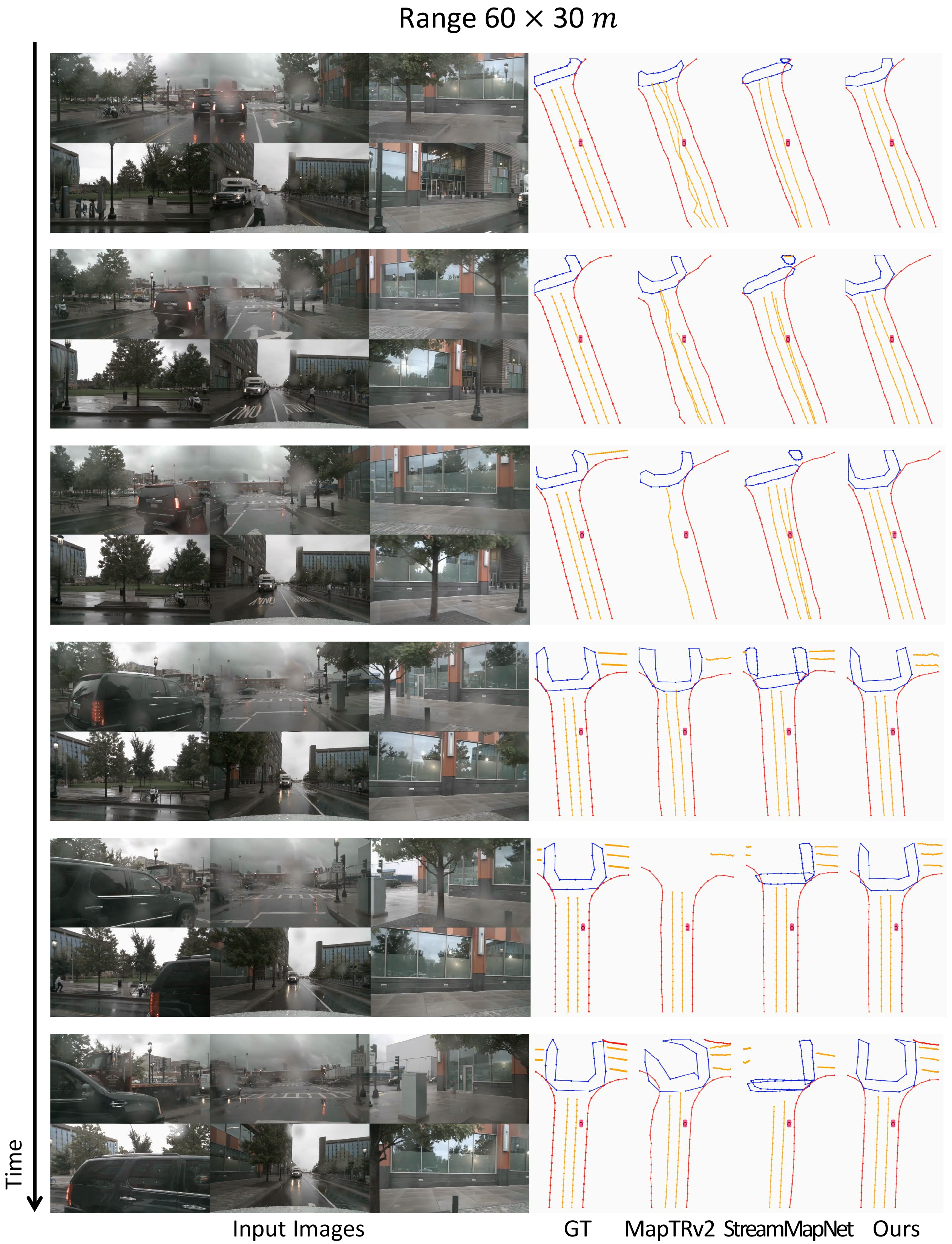}
    \caption{
    Qualitative comparisons on nuScenes \texttt{val} with 60$\times$30$m$ perception range setting.
    }
    \label{fig:fig_qualitative_appendix2}
\end{figure}

\begin{figure}
    \centering
    \includegraphics[width=.95\linewidth]{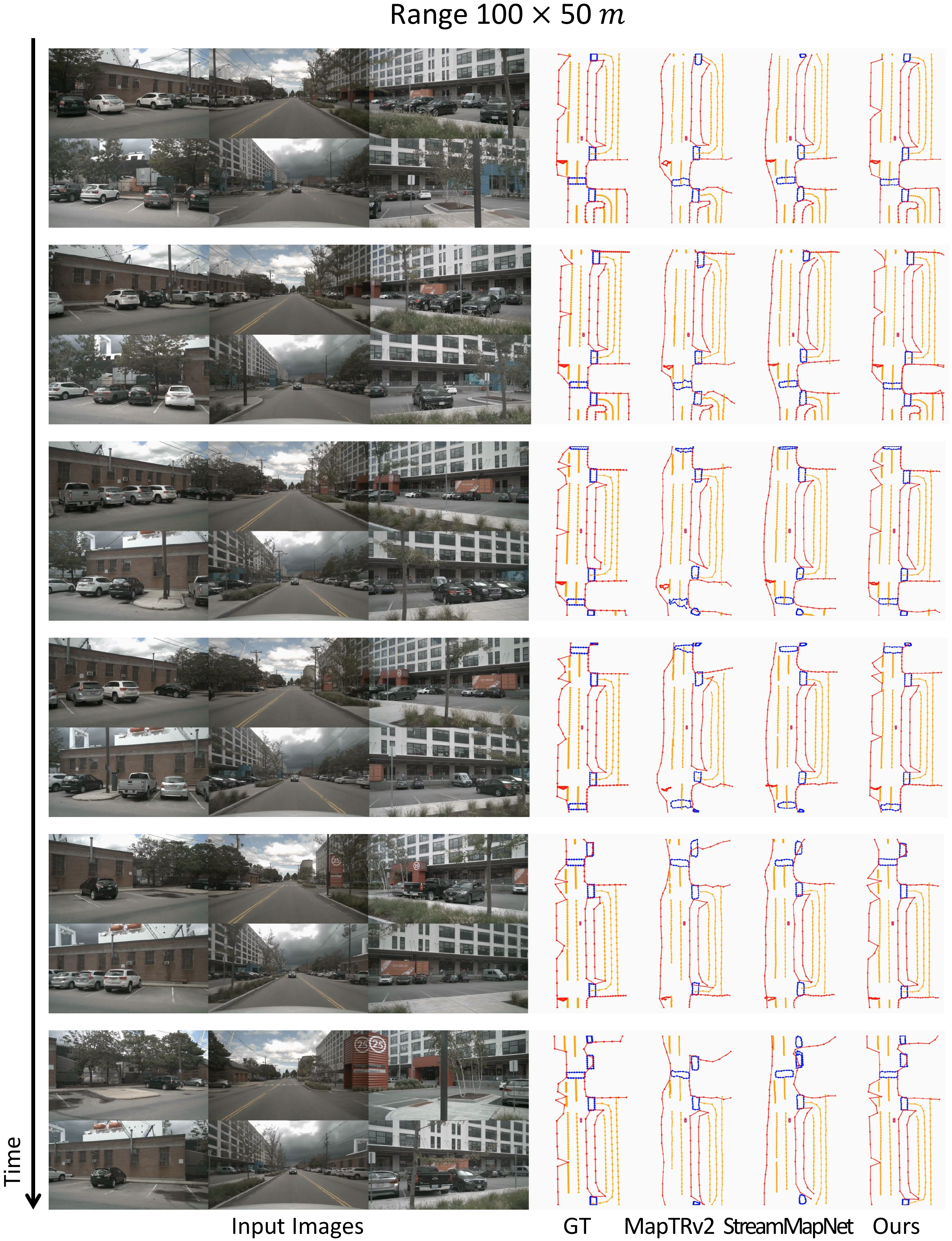}
    \caption{
    Qualitative comparisons on nuScenes \texttt{val} with 100$\times$50$m$ perception range setting.
    }
    \label{fig:fig_qualitative_appendix3}
\end{figure}

\begin{figure}
    \centering
    \includegraphics[width=.95\linewidth]{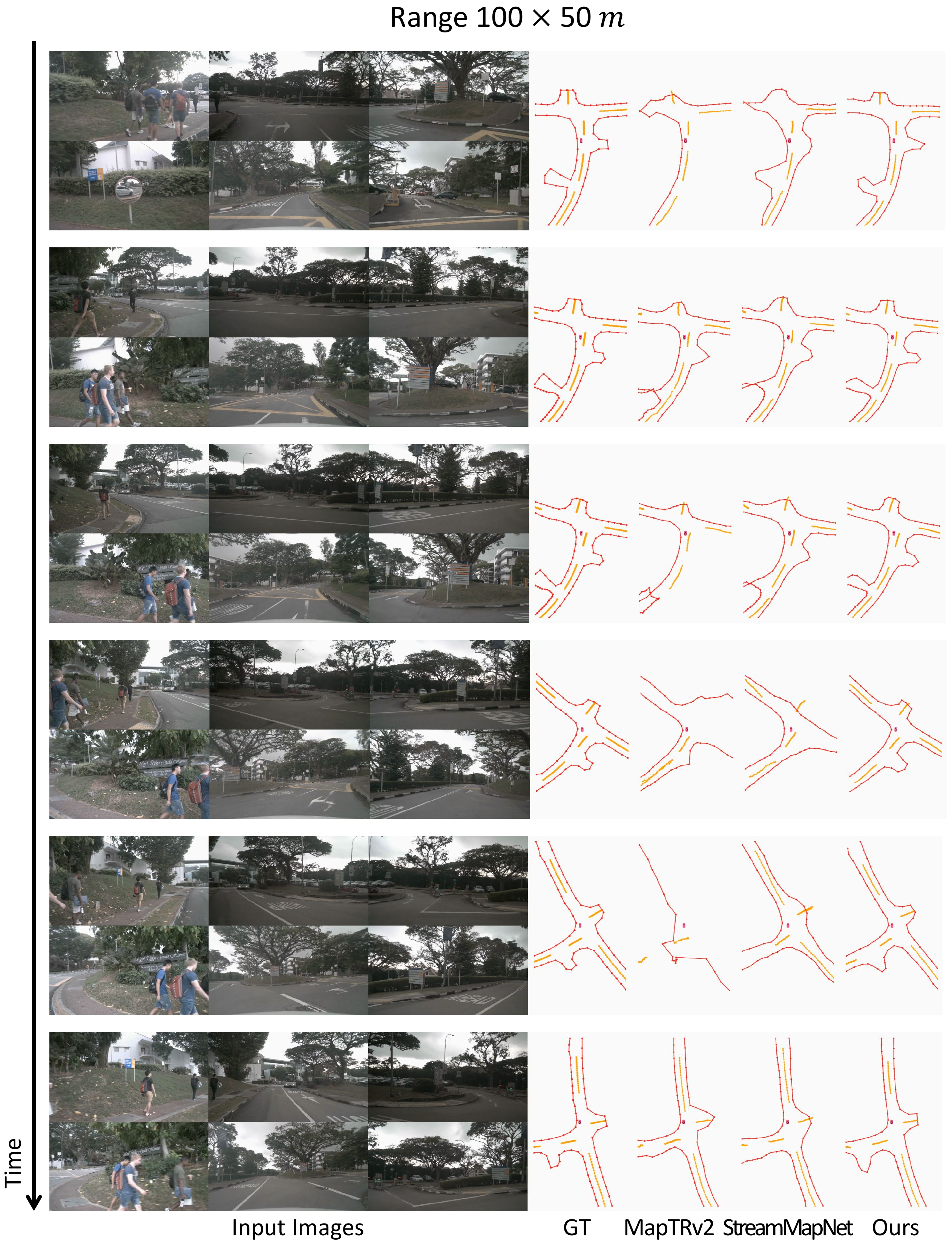}
    \caption{
    Qualitative comparisons on nuScenes \texttt{val} with 100$\times$50$m$ perception range setting.
    }
    \label{fig:fig_qualitative_appendix4}
\end{figure}

We present additional qualitative results in Figs.~\ref{fig:fig_qualitative_appendix1},~\ref{fig:fig_qualitative_appendix2},~\ref{fig:fig_qualitative_appendix3}, and~\ref{fig:fig_qualitative_appendix4}.
For all results, we provide multi-camera input and the results of MapTRv2~\cite{liao2023maptrv2}, StreamMapNet~\cite{yuan2024streammapnet}, and our \framework{}.

\subsection{Broader Impacts}
\label{border_impacts}
While our framework notably enhances the performance of online vectorized HD map construction relying on streaming multi-view camera sensors, it does not assure flawless prediction of all map elements.
Therefore, it is crucial to have a backup plan ready for safety-critical real-world applications.

\end{document}